\pdfoutput=1

\documentclass[letterpaper]{article}
\usepackage{authblk}
\usepackage[]{ACL2023}
\usepackage{longtable}
 \usepackage{booktabs}
\usepackage{times}
\usepackage{latexsym}
\usepackage[T1]{fontenc}

\usepackage[utf8]{inputenc}

\usepackage{microtype}

\usepackage{inconsolata}
\usepackage{pdflscape}
\usepackage{array}
\begin{document}

\setlength\titlebox{5cm}
%

\title{The Quest for Visual Understanding: A Journey Through the Evolution
of Visual Question Answering}




\author{Anupam Pandey$^1$, Deepjyoti Bodo$^1$, Arpan Phukan$^1$\\Asif Ekbal$^1$$^2$\\
  $^1$Department of CSE IIT Patna, $^2$School of AIDE IIT Jodhpur \\
  \texttt{ anupam\_2311cs31@iitp.ac.in, deepjyoti\_2311ai65@iitp.ac.in}\\
    \texttt{arpan\_2121cs33@iitp.ac.in, asif@iitj.ac.in/iitp.ac.in}
  \\}
  
\maketitle
\begin{abstract}
Visual Question Answering (VQA) is an interdisciplinary field that bridges the gap between computer vision (CV) and natural language processing (NLP), enabling Artificial Intelligence (AI) systems to answer questions about images. Since its inception in 2015, VQA has rapidly evolved, driven by advances in deep learning, attention mechanisms, and transformer-based models. This survey traces the journey of VQA from its early days, through major breakthroughs, such as attention mechanisms, compositional reasoning, and the rise of vision-language pre-training methods. We highlight key models, datasets, and techniques that shaped the development of VQA systems, emphasizing the pivotal role of transformer architectures and multimodal pre-training in driving recent progress. Additionally, we explore specialized applications of VQA in domains like healthcare and discuss ongoing challenges, such as dataset bias, model interpretability, and the need for common-sense reasoning. Lastly, we discuss the emerging trends in large multimodal language models and the integration of external knowledge, offering insights into the future directions of VQA. This paper aims to provide a comprehensive overview of the evolution of VQA, highlighting both its current state and potential advancements.
\end{abstract}

\section{Introduction}
Visual Question Answering (VQA) involves the development of systems that can analyze visual content and answer natural language questions about it.This seemingly simple task requires deep understanding of NLP, CV, and machine learning.It requires the integration of diverse capabilities, including image understanding, language comprehension, and multi-step reasoning.Foundational works like \cite{Fukushima1980} and \cite{krizhevsky2012imagenet} advanced CNNs for pattern recognition and object detection. \cite{girshick2014richfeaturehierarchiesaccurate} work improved segmentation, while \cite{he2015deepresiduallearningimage} introduced residual connections, enabling deeper networks for better visual content understanding. The VQA task poses significant challenges because it not only involves object detection and scene interpretation but also necessitates the ability to link the visual content with the semantic understanding of a question, often requiring nuanced contextual knowledge.

\subsection{Motivation and Importance}
The motivation for VQA stems from the broader goal
of building intelligent systems that can perceive, rea-
son, and interact with the world in ways that are
similar to human cognition. Unlike traditional im-
age recognition tasks, which focus on labeling ob-
jects in an image, VQA pushes the boundaries by ask-
ing models to answer open-ended questions—ranging
from factual queries like “What is the man holding?”
to more complex reasoning questions like ``Why is the
person smiling?''

VQA has immense potential in real-world applications, such as:

     \textbf{Healthcare:} Assisting doctors in understanding and interpreting complex medical images.
     \textbf{Education:} Enhancing interactive learning tools where students can ask questions about visual content.
     \textbf{Assistive Technologies:} Providing visually impaired users with detailed descriptions and answers about their surroundings.
    \textbf{Autonomous Systems:} Enabling robots and autonomous agents to perceive and interact with their environment based on visual queries.

\subsection{Early Beginnings of VQA}
The VQA task first gained widespread attention in
2015, when \cite{DBLP:journals/corr/AntolALMBZP15} \footnote{\url{https://visualqa.org/}} introduced VisualQA the pioneering VQA dataset and defined the task formally.
This dataset, containing hundreds of thousands of question-answer pairs tied to images, provided the first large-scale benchmark for the research community to explore the potential of multimodal AI. Before the formalization of VQA, researchers in CV and NLP worked on tasks like image captioning, object detection, and image segmentation.For instance, early work in image captioning \cite{10.5555/3045118.3045336}, object detection \cite{girshick2014richfeaturehierarchiesaccurate}, and image segmentation \cite{7298965} advanced foundational techniques such as generating descriptions, region-based object detection, and pixel-level classification, respectively.However, these efforts were limited in their ability to facilitate dynamic interactions between users and AI systems. 
VQA, on the other hand, presented an unprecedented challenge that required systems to combine object detection, relationship understanding, and reasoning capabilities.It quickly became clear that solving this task would involve addressing issues in multiple areas, such as:

     \textbf{Visual Perception:} Accurately understanding the image, including object detection, recognition, and spatial relationships.
    \textbf{Language Comprehension:} Parsing the natural language question and extracting meaningful components.
     \textbf{Reasoning:} Inferring or deducing answers, sometimes based on implicit information in the image or external knowledge.

\subsection{Objective of the Survey}
This survey aims to provide a comprehensive review of the evolution of VQA, charting its journey from its early days to the present state-of-the-art models. We explore the critical breakthroughs and innovations that have significantly advanced VQA systems, including the transition from early deep learning models \cite{krizhevsky2012imagenet} to transformer-based architectures \cite{vaswani2023attentionneed}, and the advent of vision-language pre-training \cite{lu2019vilbertpretrainingtaskagnosticvisiolinguistic} that has enabled more generalizable models.
The key areas of focus include:
\begin{itemize}
    \item Major breakthroughs in VQA architecture: The introduction of attention mechanisms, modular networks, and transformer-based models that have significantly improved performance.
    \item  The role of datasets: How benchmarks like \cite{DBLP:journals/corr/AntolALMBZP15}, \cite{johnson2017clevr}, and \cite{marino2019okvqavisualquestionanswering} have driven progress and introduced new challenges related to reasoning and common-sense knowledge.
    \item Specialized applications: Exploring how VQA has been applied in various fields such as healthcare \cite{Rajkomar2018ScalableAA} and  entertainment \cite{tapaswi2016movieqaunderstandingstoriesmovies}, where domain-specific models are making a tangible impact.
    \item Challenges and limitations: Addressing the persistent issues in VQA, such as dataset bias, model interpretability, and the need for robust reasoning over more complex question types.
\end{itemize}

\subsection{Structure of the survey}
The structure of this survey is organized chronologically to provide an overview of the key milestones and developments in Visual Question Answering (VQA) research. Each section is divided into two main paradigms: extractive and abstractive.the Section \ref{sec: section_2} reviews the early efforts and foundational work in image and language understanding that laid the groundwork for VQA. Section \ref{sec: Section_3} focuses on the deep learning era, highlighting the introduction of neural network-based models that marked the first significant leap in VQA performance. Section \ref{sec: Section_4} emphasizes the adoption of attention mechanisms and compositional reasoning, which enabled models to better understand complex relationships between objects and answer more difficult queries. In Section \ref{sec: Section_5}, we explore the revolution brought about by transformers and vision-language pre-training, which led to substantial improvements in VQA performance. Section \ref{sec: Section_6} discusses the development of specialized models for domain-specific tasks, such as medical image interpretation and region-specific VQA datasets. Section \ref{sec: Section_7} addresses current challenges and ongoing research, including efforts to overcome bias and enhance generalization. Finally, Section 8 looks ahead to the future of VQA, considering the integration of large multimodal models and the growing role of external knowledge for common-sense reasoning.

\section{Early Efforts and Foundational Work for VQA}\label{sec: section_2}
Before VQA emerged as a formalized task, advancements in image understanding and language modeling laid the foundation for integrating vision and language. Early works in object detection \cite{990517}, scene recognition \cite{Xiao2010SUN}, and NLP innovations such as word embeddings \cite{Mikolov2013Efficient} and sequence modeling \cite{sutskever2014sequencesequencelearningneural} were instrumental in bridging these domains. Object detection and scene recognition enabled visual feature extraction, while language modeling advancements, including RNNs \cite{10.1162/neco.1997.9.8.1735} and transformers \cite{vaswani2023attentionneed}, improved textual representations.

This convergence of vision and language technologies catalyzed the development of tasks that require both modalities, such as image captioning \cite{10.5555/3045118.3045336}  and visual dialogue \cite{das2017visualdialog}, setting the stage for the evolution of VQA. In this section, we outline these key advancements, categorizing them into extractive and abstractive paradigms, which significantly influenced the subsequent development of VQA systems.

\subsection{The role of Image Understanding (Object Detection and Scene Recognition) in VQA}Advancements in image understanding began with the introduction of AlexNet \cite{NIPS2012_c399862d} , which employed convolutional neural networks (CNNs) to achieve state-of-the-art performance on the large-scale ImageNet dataset \cite{deng2009imagenet}. AlexNet’s ability to classify objects into a predefined set of labels demonstrated the power of deep learning for extracting visual features, making it a cornerstone for early VQA systems. Building on this, R-CNN \cite{girshick2014rich} extended this capability to fine-grained region-based object detection, enabling models to localize specific areas within an image. These advancements were instrumental for extractive VQA, where models needed to identify predefined objects or answer simple queries like “What is in the top-left corner?'' Meanwhile, datasets like PASCAL VOC \cite{everingham2010pascal} provided annotated benchmarks for object classification and segmentation, further driving progress in \textbf{extractive} tasks.

In contrast, \textbf{abstractive} approaches in image understanding moved beyond object-level recognition to describe the entire scenes. The large-scale SUN Database \cite{Xiao2010SUN} pioneered scene categorization, enabling models to interpret the overall context of an image rather than isolated objects. While ImageNet \cite{deng2009imagenet} was primarily designed for object classification, its diversity inspired future models to explore richer scene-level abstractions, supporting tasks that required reasoning about the broader context, such as answering “What is happening in this image?” These contributions laid the groundwork for models that could synthesize free-form descriptions, enabling abstractive reasoning in VQA.

\subsection{The Role of Natural Language Processing in VQA}
Simultaneously, significant strides in NLP provided the textual modeling capabilities necessary for VQA. The introduction of Word2Vec \cite{mikolov2013efficientestimationwordrepresentations} marked a breakthrough in embedding words as continuous vectors, capturing semantic relationships like synonymy and analogy. These embeddings enabled extractive VQA systems to map the predefined question patterns to specific answers, such as linking “What color is the ball?” to “Red.” Earlier works like the Penn Treebank dataset \cite{marcus1993building} provided syntactic structures for language parsing, offering insights into grammatical relationships, which proved useful for models handling structured question-answer tasks. These works formalized extractive NLP methods, emphasizing predefined mappings between the inputs and outputs.

On the other hand, abstractive NLP approaches began exploring dynamic text generation, enabling free-form responses. Recurrent Neural Networks (RNNs), introduced by \cite{elman1990finding}, provided the first framework for processing sequential data, allowing models to generate text dynamically based on context. This capability evolved into encoder-decoder architectures, such as the attention-based model by \cite{bahdanau2016neuralmachinetranslationjointly}, which introduced alignment mechanisms in sequence-to-sequence tasks. Similarly, \cite{cho2014learningphraserepresentationsusing} proposed the GRU-based encoder-decoder model, offering computationally efficient alternatives to long short-term memory (LSTM) \cite{article} networks.In VQA, encoder-decoder frameworks inspired abstractive models to handle questions requiring descriptive reasoning or multi-step logic. For instance, \cite{Andreas2016Neural} introduced neural module networks (NMNs) that dynamically compose task-specific reasoning modules for multi-step logic. Additionally, \cite{johnson2017clevr} proposed the CLEVR dataset and model, focusing on compositional reasoning to answer complex visual queries. These advances significantly influenced the design of VQA systems, particularly in generating detailed and contextually relevant answers.

\subsection{Bridging Vision and Language}
Efforts to bridge vision and language began in earnest with tasks like image captioning, where models generated textual descriptions of visual content. In the extractive paradigm, works like \cite{karpathy2015deep} aligned image regions with caption fragments, enabling models to match predefined visual features with corresponding text. Similarly, datasets like MS COCO \cite{lin2014microsoft} provided annotated captions for images, supporting fixed-answer approaches to describing visual content. These early efforts inspired extractive VQA, where models focused on identifying regions or objects to provide predefined responses.

Meanwhile, the abstractive paradigm in image captioning demonstrated the potential for generating free-form, context-aware descriptions. Show and Tell, introduced by \cite{vinyals2015show} combined CNNs for image encoding and RNNs for text generation to produce captions like “A dog sitting on the grass.” This work was pivotal in showing that models could synthesize descriptive answers rather than relying on fixed outputs. Shortly thereafter, Show, Attend and Tell by\cite{10.5555/3045118.3045336} integrated attention mechanisms into this framework, enabling the model to focus on specific image regions while generating captions, paving the way for attention-based abstractive VQA.

\subsection{Early Attempts at Visual Question Answering}
Early formalizations of VQA began with the DAQUAR dataset by \cite{Malinowski2014Multimodal}, which introduced visual question-answering tasks focused on indoor scenes.These tasks were primarily extractive, with predefined answers requiring models to identify objects or attributes.The Visual Turing Test \cite{doi:10.1073/pnas.1422953112} proposed in the same year formalized the evaluation of vision-language reasoning, emphasizing the ability of systems to match predefined outputs to given inputs.In contrast, datasets like CLEVR \cite{johnson2017clevr} pushed the boundaries of abstractive VQA, requiring multi-step reasoning and free-form answers to compositional questions such as “What is the color of the sphere to the left of the cube?”.By the mid-2010s, these advancements converged to formalize VQA as a distinct research area.The VQA Dataset v1.0, introduced by \cite{Antol2015VQA}, marked a turning point, providing a large-scale benchmark that included both multiple-choice and open-ended questions.Although the data set supported both paradigms, the early models relied mainly on extractive techniques, combining CNNs for image encoding and LSTMs for question processing to classify answers from a fixed pool. This milestone formalized VQA, providing a foundation for the rapid advancements that followed.

\section{Early Deep Learning Models and Representation Fusion}\label{sec: Section_3}
VQA's early deep learning era was characterized by advancements in multimodal fusion and representation learning.During this time, models that approached the job using extractive paradigms—which select predefined answers—and abstractive paradigms—which produce open-ended textual answers—began to appear.Integrating text and visual representations into cohesive frameworks—often using deep neural networks—was essential to these advancements. In the parts that follow, we go into further detail about:

\subsection{CNN-LSTM Models: The First Generation of VQA Models}
Extractive Paradigm:
The first generation of Visual Question Answering (VQA) models utilized CNNs for image feature extraction and LSTMs for processing questions. \cite{Antol2015VQA} formalized VQA with the introduction of the VQA v1.0 dataset, where CNN-LSTM models became the default approach. For example, image features were extracted using pre-trained CNNs like AlexNet \cite{NIPS2012_c399862d} and VGGNet \cite{simonyan2015deepconvolutionalnetworkslargescale}, which provided global image representations. The questions were encoded using LSTMs, which transformed the text into a fixed-length vector. These two representations were then fused, typically through concatenation or element-wise operations, followed by a classifier predicting answers from a fixed vocabulary. An early example of this approach is found in \cite{Malinowski2014Multimodal}, where models were designed to answer object-centric questions such as, "What is the object in the center?" The model would analyze the image and correctly identify that the object in the center is a "bottle." These early models demonstrated the potential of deep learning for extractive reasoning, but were limited by their reliance on simple feature fusion techniques that did not capture complex relationships within the visual content.

Abstractive Paradigm:
Abstractive systems during this era adapted CNN-LSTM models to generate open-ended answers. For example, \cite{Gao2015Are} proposed an encoder-decoder framework where visual features extracted by a CNN were passed to an RNN for dynamic text generation. Given a question like “What is the man doing?”, these models generated diverse answers such as “A man is riding a horse”, rather than selecting from predefined categories.
These methods drew inspiration from advancements in image captioning \cite{vinyals2015show}, leveraging free-form text generation techniques to respond to visual questions. However, the generated responses often suffered from a lack of fine-grained multimodal reasoning, leading to generic or overly simplistic answers that failed to fully capture complex visual and linguistic nuances.

\subsection{Bilinear Pooling: Improving Multimodal Fusion for VQA}
Extractive Paradigm:
The introduction of bilinear pooling techniques marked a significant improvement in multimodal feature fusion by capturing fine-grained interactions between visual and textual features. \cite{Fukui2016Multimodal} proposed Multimodal Compact Bilinear (MCB) pooling, which approximated full bilinear interactions using compact representations. This approach enhanced the model's ability to reason about visual details, enabling it to excel in tasks such as attribute-based questions like “What color is the ball?”, with accurate responses such as “The color of the ball is Red”.Building on this, \cite{Kim2017Bilinear} introduced Multimodal Low-rank Bilinear (MLB) pooling, which reduced computational complexity while maintaining high performance. These techniques became particularly effective for questions requiring detailed attribute reasoning.

Abstractive Paradigm:
Abstractive models also benefited from bilinear pooling, which enabled the generation of more context-aware and nuanced responses. For instance, \cite{Kim2018Learning} extended bilinear pooling to generative frameworks, allowing the model to answer complex questions such as “What is happening in this image?” with responses like “A child is playing with a ball on the beach.”Bilinear pooling improved cross-modal interactions by learning joint feature representations that captured intricate dependencies between visual elements (e.g., object relationships, spatial configurations) and linguistic context. This richer representation facilitated logical reasoning, enabling abstractive models to dynamically generate contextually appropriate answers. By leveraging these interactions, abstractive systems could reason beyond simple feature alignment, accounting for complex scene details and generating responses that better mirrored human-level understanding.

\subsection{Attention Mechanisms: Learning to Focus}
Extractive Paradigm:
The introduction of attention mechanisms allowed models to dynamically focus on relevant image regions based on the question context. \cite{Yang2016Stacked} proposed the Stacked Attention Network (SAN), which iteratively refined attention over the image to identify the most relevant regions. For example, it excelled in answering questions like “What is to the left of the car?”
Attention improved extractive VQA by reducing reliance on global features, enabling models to reason about specific regions.
Building on this idea, \cite{lu2016hierarchical} introduced the Hierarchical Co-Attention Network, where question-guided attention was utilized to compute attention weights for both image regions and the question embedding. This allowed VQA models to explicitly link visual features with linguistic context, achieving better interpretability and accuracy in responses.
Subsequent advancements, such as Dual Attention Networks (DAN) proposed by \cite{nam2017dual}, modeled both visual and textual attention jointly. By attending to critical parts of both modalities simultaneously, these models demonstrated strong performance in answering questions that required a fine-grained understanding of spatial and semantic relationships, such as "What is in front of the table? Ans. Chair"
Another notable contribution, the Spatial Memory Network (SMN) by \cite{chen2017spatial}, leveraged attention to iteratively refine spatial reasoning. This approach proved particularly effective for questions that required understanding precise spatial relationships between objects, such as answering "What is between the cup and the laptop? Ans. Notepad"
These early advancements in extractive attention mechanisms established a strong foundation for more sophisticated approaches. They underscored the importance of focusing on specific regions of the image, rather than relying solely on global features, thereby enabling models to reason over visual content with improved precision and interpretability.

Abstractive Paradigm:
Attention mechanisms also played a pivotal role in abstractive VQA, enabling the generation of coherent and contextually accurate answers. The Show, Attend, and Tell framework by \cite{10.5555/3045118.3045336}, originally designed for image captioning, demonstrated how attention could guide dynamic text generation by focusing on relevant visual areas. Adapted for VQA, these techniques enabled models to answer questions such as “What is the dog doing?” with responses like “The dog is chasing a ball in a park”, incorporating both visual and textual context.

Beyond this, \cite{lu2017knowinglookadaptiveattention} proposed a mixed attention mechanism that combined visual and linguistic reasoning, allowing the model to dynamically balance image-specific and question-specific focus. Similarly, \cite{zhou2019unifiedvisionlanguagepretrainingimage} extended attention to transformer-based architectures, integrating multi-headed attention for joint reasoning across vision and language tasks. This approach facilitated abstractive answers by leveraging contextual dependencies across both modalities.By synthesizing these advancements, attention mechanisms enabled abstractive systems to dynamically generate nuanced answers, leveraging fine-grained cross-modal interactions to interpret and describe complex scenes. This integration significantly improved the ability of VQA systems to reason contextually and provide human-like responses.

\subsection{Bottom-Up and Top-Down Attention: A Major Breakthrough}
Extractive Paradigm:
\cite{Anderson2018Bottom} introduced the bottom-up and top-down attention framework, which significantly advanced VQA. This model used Faster R-CNN \cite{girshick2014richfeaturehierarchiesaccurate} to extract object-level features (bottom-up attention) and applied a question-driven focus mechanism (top-down attention) to reason about relationships. This approach achieved state-of-the-art results on extractive tasks, such as identifying objects or answering positional queries.

Abstractive Paradigm:
In the abstractive paradigm, the bottom-up and top-down attention framework enabled richer relational reasoning, allowing for more dynamic and descriptive answers. For instance, given the question “What is happening in the image?” models could generate open-ended answers such as “A person is sitting under a tree while reading a book,” by focusing on the interactions between specific objects, their attributes, and contextual relationships. This approach not only improved the generation of descriptive answers but also facilitated multi-step reasoning, \cite{Tan2019LXMERT} integrated both vision and language through cross-modal attention, enhancing the model’s ability to generate complex, reasoning-based answers in the abstractive setting.

\subsection{The Shift towards Advanced Models for VQA}
Extractive Paradigm:
With the rise of pre-trained architectures like UNITER \cite{Chen2020UNITER} and VisualBERT \cite{Li2019VisualBERT}, multimodal fusion evolved into end-to-end systems pre-trained on large datasets. These transformer-based models surpassed traditional CNN-LSTM architectures by enabling finer-grained feature extraction and reasoning for extractive tasks. They achieved breakthroughs in benchmarks like VQA v2.0 \cite{Goyal2017Visual} by leveraging large-scale pre-training on multimodal data.
Abstractive Paradigm:
For abstractive reasoning, transformers like Oscar \cite{Li2020Oscar} demonstrated the ability to generate rich, contextually informed answers. These models integrated multimodal embeddings directly into the decoding process, enabling seamless generation of open-ended answers.

\section{Compositional Reasoning and Modular Networks For VQA}\label{sec: Section_4}
As VQA tasks grew more complex, requiring reasoning about multiple objects, relationships, and attributes, traditional neural network approaches began to face limitations. This motivated the development of compositional reasoning models, which broke down complex queries into smaller, interpretable steps. Neural Module Networks (NMNs) \cite{Andreas2016Neural} introduced task-specific modules for compositional reasoning, while graph-based approaches leveraged structured representations like scene graphs to encode relationships between objects. Later, Compositional Attention Networks (MAC) \cite{Hudson2018Compositional} introduced iterative reasoning mechanisms, and pre-trained transformers signaled the decline of modular networks by performing implicit reasoning over multimodal data.

This section categorizes compositional reasoning approaches into extractive and abstractive paradigms, explaining their significance, methodologies, and limitations.

\subsection{Neural Module Networks (NMNs): A Modular Approach to Reasoning}
Extractive Paradigm:
NMNs \cite{Andreas2016Neural} represented a significant shift in VQA by introducing dynamically assembled networks tailored to the structure of a query. These models were built using small, reusable modules (e.g., Locate Tree, Find Left, Recognize Color), each responsible for a specific task. For example, to answer the query “What is the color of the car to the left of the tree?”, NMNs would:
Use a Locate Tree module to identify the tree in the image.
Deploy a Find Left module to locate objects to the left of the tree.
Apply a Recognize Color module to determine the car’s color.
This explicit, modular approach allowed NMNs to break down complex questions into logical sub-tasks, making them highly effective for extractive VQA, where answers were derived from well-defined visual attributes or spatial relationships. However, NMNs depended heavily on the accurate parsing of queries and reliable object detection, which posed challenges in ambiguous or noisy scenarios. Moreover, their reliance on task-specific modules made them computationally expensive and less generalizable to diverse datasets.

Abstractive Paradigm:
To extend beyond predefined answers, Dynamic Neural Module Networks \cite{Johnson2017Inferring} introduced models capable of adapting their structure dynamically based on a query’s semantics. For example, a model could synthesize reasoning steps for a question like “Describe the activities in the scene” by assembling modules for detecting objects, identifying interactions, and synthesizing a free-form description. These abstractive NMNs could reason over complex relationships and provide open-ended answers. However, the computational cost of constructing dynamic networks and challenges in handling ambiguous language limited their scalability for real-world applications.

\subsection{Graph-Based Visual Question Answering}
Extractive Paradigm:
Graph-based approaches brought structured representations into VQA, leveraging scene graphs \cite{Krishna2017Visual} and Graph Neural Networks (GNNs) \cite{Teney2017Graph} to encode relationships between objects. A scene graph represents objects (e.g., "tree," "car") as nodes and their relationships (e.g., "to the left of," "under") as edges. For example, to answer the query “What is next to the tree?”, a scene graph would: Identify all objects in the image as nodes (e.g., "tree," "car," "house").
Traverse edges labeled with spatial relationships (e.g., "next to") to find the relevant object ("car").
Graph-based reasoning provided interpretable pathways for extractive VQA tasks, enabling precise alignment between visual elements and textual queries. These methods excelled in tasks requiring explicit reasoning over spatial or relational data. However, they relied heavily on accurate object detection and relationship extraction, which limited their robustness in real-world, noisy environments.

Abstractive Paradigm:
For abstractive reasoning, dynamic scene graphs \cite{Yang2018Dynamic} and knowledge-augmented networks \cite{Narasimhan2022Knowledge} extended graph-based methods by integrating external knowledge and contextual information. For example, answering “What might the person be doing under the tree?” involved not just detecting objects and relationships but also inferring activities based on scene context and external knowledge. These models synthesized context-aware, free-form answers by reasoning over visual elements and incorporating prior knowledge.While these abstractive methods expanded the scope of graph-based reasoning, challenges like the high cost of constructing large-scale scene graphs and the reliance on accurate annotations limited their widespread adoption.

\subsection{Compositional Attention Networks (MAC) for VQA}
Extractive Paradigm:
The Compositional Attention Network (MAC)\cite{Hudson2018Compositional} introduced an innovative approach to reasoning by employing a recurrent attention mechanism that iteratively refined its understanding of visual and textual inputs. Each reasoning step in the MAC model was guided by a structured memory unit and attention mechanism, enabling the model to perform sequential reasoning for complex queries. For example, to answer “What is the color of the car in front of the house?”, the MAC model would:
Attend to the house.
Locate the car in front of the house.
Recognize the car’s color.
This step-by-step process allowed MAC to excel in extractive reasoning tasks, providing interpretable reasoning pathways and outperforming traditional attention mechanisms. However, its reliance on iterative attention made it computationally expensive, and it sometimes struggled with ambiguous or noisy queries.

Abstractive Paradigm:
MAC networks were extended to abstractive tasks by incorporating neural-symbolic reasoning \cite{Yi2018Neural}, where logical operations were performed over explicit visual representations like scene graphs. For example, to answer “Describe the scene,” the model could sequentially identify objects, relationships, and activities to synthesize a coherent narrative.These abstractive extensions demonstrated impressive capabilities for multi-step reasoning and complex queries. However, the computational cost of iterative reasoning and the difficulty of integrating symbolic reasoning with neural networks remained significant bottlenecks.

\subsection{The Rise of Pre-trained Transformers and the decline of Modular Networks}
Extractive Paradigm:
Pre-trained transformer models like LXMERT \cite{Tan2019LXMERT} and VilBERT \cite{lu2019vilbert} marked a paradigm shift in VQA by learning joint image-text representations through self-attention mechanisms. These models implicitly captured relationships and compositional structures, eliminating the need for explicit modular or graph-based architectures. For example, a question like “What is the color of the object to the left of the tree?” could be answered by leveraging the global attention mechanism to focus on relevant image regions and text tokens. Transformers excelled in extractive tasks due to their ability to process multimodal inputs end-to-end, achieving state-of-the-art results without requiring handcrafted modules or explicit reasoning graphs.
Abstractive Paradigm:
For abstractive tasks, transformers like UNITER \cite{Chen2020UNITER} and OSCAR  \cite{Li2020Oscar} integrated external knowledge and context to generate rich, free-form answers. By leveraging large-scale pre-training on multimodal datasets, these models synthesized detailed, context-aware descriptions, addressing queries like “Describe the interaction between the people in the image.”Transformers' end-to-end learning approach and ability to generalize across diverse datasets signaled the decline of modular networks. However, the trade-offs included a lack of interpretability and the requirement for significant computational resources during pre-training and inference.

\section{Transformer Models and Vision-Language Pre-training}\label{sec: Section_5}
Transformer-based models have revolutionized Vision-Question Answering (VQA) by introducing advanced mechanisms to align and process visual and textual inputs. These models utilize the self-attention mechanism inherent in transformers, which enables the models to capture intricate relationships between image content and textual queries. In VQA tasks, some models focus on extractive approaches, aiming to identify specific regions or objects in an image that directly relate to the question posed. Such models, like LXMERT \cite{Tan2019LXMERT} and VilBERT  \cite{lu2019vilbertpretrainingtaskagnosticvisiolinguistic}, employ cross-modal attention mechanisms to align visual features with the textual query. LXMERT connects objects in an image with their corresponding linguistic elements, excelling in spatial and relational tasks like "What is the person holding?" . Meanwhile, VilBERT strengthens this process through bilateral cross-modal interactions, enhancing its ability to reason about detailed object relationships. These extractive models excel at pinpointing specific visual elements necessary for answering precise questions.

In contrast, other transformer-based models adopt an abstractive approach to VQA, where the goal is to generate descriptive or inferential answers rather than simply extracting specific image regions. Models such as CLIP \cite{radford2021clip} and ViLT \cite{radford2021clip} utilize a broader understanding of the relationships between visual and textual content to answer more general, abstract questions. CLIP, pretrained on large datasets of image-text pairs, allows for zero-shot VQA, answering complex queries about context or emotions without requiring task-specific fine-tuning CLIP. Similarly, ViLT processes raw image patches and text tokens through a unified transformer architecture, bypassing traditional region-based supervision to offer efficient and holistic reasoning ViLT. These abstractive models perform well on tasks that require understanding of context, such as answering questions like "Why is the person running?" or "What is the emotion of the person?" Both extractive and abstractive models showcase how transformer-based architectures are reshaping VQA, offering both precise extraction and nuanced generation of answers.

\subsection{Cross-Modal Alignment and Feature Integration in VQA}
Extractive approaches in VQA emphasize the precise alignment between visual and textual modalities to generate accurate responses. These methods often rely on learning shared representations of input modalities to facilitate cross-modal reasoning. For instance, UNITER \cite{Chen2020UNITER} achieves universal image-text representation learning by leveraging both vision and language tasks within a unified framework, effectively aligning image regions and textual embeddings. Similarly, Oscar \cite{Li2020Oscar} enhances visual grounding by incorporating object tags as anchor points during training, significantly improving the alignment between visual features and textual queries. These extractive models are benchmarked on datasets like \cite{hudson2019gqanewdatasetrealworld}, \cite{suhr2019corpusreasoningnaturallanguage}, and \cite{Goyal2017Visual}, which provide diverse and challenging scenarios for evaluating fine-grained question answering capabilities. Despite their success, these methods heavily depend on pre-extracted region-based visual features and often struggle to generalize to more dynamic and open-ended contexts.

Abstractive approaches, on the other hand, extend beyond simple alignment and integrate features to generate comprehensive and contextually enriched answers. These methods aim to synthesize new information by abstracting visual and textual inputs, going beyond direct retrieval to provide more nuanced responses. Recent advancements in multimodal learning have explored transformer-based architectures capable of leveraging end-to-end learning without relying on pre-extracted region features. While abstractive paradigms offer greater flexibility and potential for broader contextual understanding, their efficacy is hindered by limitations in multimodal representation learning and the lack of datasets specifically designed to evaluate such capabilities comprehensively. Future advancements in datasets and model architectures are essential to overcome these challenges and realize the full potential of abstractive VQA systems.

\subsection{Unified Understanding and Generalization in VQA}

Extractive approaches in VQA prioritize retrieving precise answers by aligning visual inputs and textual queries. These models employ techniques that map both modalities into a shared representational space to facilitate accurate reasoning. For example, UNITER \cite{Chen2020UNITER} and Oscar \cite{Li2020Oscar} have demonstrated significant advancements in aligning textual embeddings with visual regions, leveraging datasets such as \cite{hudson2019gqanewdatasetrealworld}, \cite{suhr2019corpusreasoningnaturallanguage}, and \cite{Goyal2017Visual} to fine-tune their performance. While these extractive models excel in structured and straightforward tasks, they often falter in open-ended or dynamic scenarios, highlighting a limitation in their ability to generalize.

Abstractive approaches in VQA are characterized by their ability to generate descriptive or inferential answers, moving beyond simple retrieval to synthesize contextually rich responses. These methods leverage advancements in holistic reasoning to integrate vision and language more effectively. For example, Flamingo \cite{alayrac2022flamingovisuallanguagemodel} combines vision and language pretraining to enable few-shot learning across multimodal tasks, demonstrating exceptional flexibility and adaptability. Similarly, Blip-2 \cite{li2023blip2bootstrappinglanguageimagepretraining} incorporates a lightweight Q-Former module and frozen vision-language encoders to enhance efficiency in abstractive reasoning, enabling robust performance without sacrificing computational efficiency. These innovations exemplify how abstractive models are reshaping the landscape of multimodal understanding.

The evolution of abstractive approaches also extends to dynamic visual content through video-based setups. Models, such as VideoBERT \cite{sun2019videobertjointmodelvideo} and Frozen in Time \cite{bain2022frozentimejointvideo} effectively process sequential visual information, allowing them to handle temporal dependencies and provide contextually enriched answers for video-based VQA tasks. Benchmarks like TDIUC \cite{kafle2017analysisvisualquestionanswering}, TextVQA \cite{singh2019vqamodelsread}, and VizWiz \cite{gurari2018vizwizgrandchallengeanswering} play a critical role in evaluating these systems, offering diverse scenarios that test their abstractive capabilities. However, challenges remain, particularly in understanding abstract reasoning and navigating domain-specific contexts. Addressing these limitations will require more comprehensive datasets and advancements in model architectures to further enhance the generalization and holistic reasoning of abstractive VQA systems.

\section{Domain-Specific VQA Models and Real-World Applications}\label{sec: Section_6}
As Visual Question Answering (VQA) models evolved with the advent of transformer-based architectures and vision-language pre-training, researchers began exploring their utility in domain-specific contexts. These applications require VQA systems to operate effectively on specialized datasets, address complex tasks, and deliver high levels of accuracy. Fields like healthcare, education, security, and entertainment have emerged as key areas where VQA can extend beyond general vision-language tasks to provide meaningful real-world solutions.
This section delves into the applications of VQA in domain-specific settings, with an emphasis on medical and entertainment-based VQA systems. It highlights prominent datasets, key models, and the unique challenges encountered in deploying VQA systems in these specialized fields.
\subsection{Medical VQA: Enhancing Diagnostics and Interpretation}

In the medical domain, extractive VQA systems emphasize accurate identification of specific elements within medical images, such as detecting abnormalities, identifying organs, or determining imaging modalities. Datasets like VQA-Med \cite{ImageCLEFVQA-Med2018} and VQA-RAD \cite{lau2018vqa_rad} provide essential benchmarks, offering radiological images paired with clinical questions. These datasets have facilitated the development of systems like MMBERT \cite{khare2021mmbert}, which leverages multimodal BERT pretraining to enhance precision in identifying conditions, such as fractures or pneumonia. Similarly, \cite{Moon_2022} investigated vision-language pretraining using deep transformer architectures for chest X-ray datasets, demonstrating significant improvements in extracting diagnostic information. These systems prioritize interpretability and clinical relevance, ensuring that outputs are actionable in medical decision-making.

Abstractive medical VQA goes beyond explicit identification tasks, aiming to generate detailed, context-aware answers that combine visual data with external domain knowledge. For example, \cite{li2024systematicevaluationgpt4vsmultimodal} analyzed the multimodal capabilities of GPT-4V \cite{Radford2018ImprovingLU} in synthesizing information from medical images and clinical guidelines to address complex diagnostic queries, such as identifying diseases based on symptoms. Models, such as \cite{bioengineering10030380} benchmarked vision-language frameworks on datasets like \cite{lau2018vqa_rad}, integrating multimodal transformers to generate nuanced responses that link imaging findings with broader clinical contexts. Despite challenges such as variability in datasets and the integration of domain expertise, abstractive VQA models showcase transformative potential for enhancing diagnostics and improving clinical workflows.

\subsection{Entertainment or Movie-VQA
}
In entertainment, extractive VQA focuses on directly retrieving information from video content, such as identifying characters, actions, or key dialogues. Datasets like MovieQA \cite{tapaswi2016movieqaunderstandingstoriesmovies} and Social-IQ \cite{Zadeh_2019_CVPR} provide structured benchmarks for such tasks. Models like STAGE \cite{cherian2022251dspatiotemporalscenegraphs} utilize scene graphs to align visual and textual elements, supporting tasks like tracking objects across scenes and resolving relationships between characters. These extractive approaches enable applications like video summarization and recommendation systems.

Abstractive VQA in the entertainment domain generates high-level narrative insights, requiring advanced reasoning over temporal dependencies and multimodal inputs. For example, models like MMFT \cite{khan2020mmftbertmultimodalfusiontransformer} and  
HME \cite{chandar2016hierarchicalmemorynetworks} infer plot developments, character motivations, or thematic elements by synthesizing diverse inputs such as audio, text, and visuals. These abstractive approaches enhance interactive media and accessibility solutions but face challenges like handling cultural nuances, genre-specific biases, and subjectivity in interpretation.
\subsection{Applications in Other Specialized Domains}
Domain-specific Visual Question Answering (VQA) spans diverse fields, including fashion, sports, finance, scientific research, and fine-grained data analysis, with specialized datasets and models driving advancements. In fashion, FashionVQA \cite{wang2022fashionvqadomainspecificvisualquestion} comprises over 200,000 image-question pairs about attributes like color and material, leveraging models such as MCAN \cite{yu2019deepmodularcoattentionnetworks} and MUTAN \cite{benyounes2017mutanmultimodaltuckerfusion} to achieve state-of-the-art performance.For sports, Sports-QA \cite{li2024sportsqalargescalevideoquestion} enables reasoning over professional sports videos, offering fine-grained temporal annotations for activities like basketball and gymnastics.In scientific domains, datasets like PlotQA \cite{methani2020plotqareasoningscientificplots} facilitate reasoning over charts and plots, answering quantitative questions by combining visual and textual analysis. Another example is FigureQA \cite{kahou2018figureqaannotatedfiguredataset}, which focuses on graphical elements in scientific figures to support VQA tasks
. Fine-grained datasets like Visual Genome \cite{krishna2016visualgenomeconnectinglanguage} also contribute to complex reasoning across various structured and unstructured domains. These datasets and models showcase the versatility of VQA in addressing real-world problems across specialized sectors.

\section{Challenges and Ongoing Research}\label{sec: Section_7}
Video Question Answering (VQA) faces significant challenges, with temporal reasoning being a critical hurdle. It involves identifying event sequences and understanding causality in videos, demanding models capable of effectively processing these relationships. Multimodal integration presents another obstacle, as VQA systems must align visual, textual, and auditory data using advanced attention mechanisms to ensure coherence across modalities \cite{li2024surveybenchmarksmultimodallarge}. Limited availability of high-quality annotated datasets, particularly in niche domains or non-English contexts, further complicates the development and evaluation of robust models. Scalability poses a concern, with the computational demands of processing long-form video content often exceeding current hardware and algorithmic capabilities. Additionally, addressing open-ended questions requires sophisticated generative models capable of producing detailed and nuanced answers, pushing VQA systems beyond simple classification tasks. Ensuring generalization across diverse video domains and addressing biases in the data and models also remain significant challenges for the field.
The evolution of VQA systems depends on resolving these foundational issues. Effective temporal modeling requires advanced techniques for understanding event causality and sequence prediction. Seamless multimodal integration demands robust architectures capable of synthesizing diverse data inputs. Efforts must also focus on overcoming ambiguity in natural language to enhance model adaptability, particularly in resource-scarce and linguistically diverse environments. Addressing data scarcity will require innovations in data augmentation and self-supervised learning methods to support training and evaluation comprehensively.

Resource efficiency is another critical area, as long-form video processing necessitates scalable and computationally efficient architectures. Enhancing generative capabilities is vital for tackling open-ended queries and expanding the scope of VQA applications. Finally, prioritizing ethical considerations and ensuring fairness and generalization across varied domains will be essential to developing reliable and inclusive systems. Future advancements should focus on integrating causal reasoning, improving computational efficiency, and fostering inclusivity to achieve robust and high-performing VQA solutions.

\section{Future Directions}\label{sec: Section_8}
The field of Video Question Answering (VQA) presents vast potential for growth, with numerous opportunities for future research across various dimensions, including tasks, domains, languages, and algorithmic advancements. Future research can explore novel tasks requiring deeper temporal and contextual understanding, such as complex activity recognition, causal reasoning, and real-time VQA applications for video surveillance, autonomous systems, and emergency response \cite{zhong2022videoqa}. Specialized domains like education, medical imaging, and legal evidence analysis can also benefit from VQA systems tailored to their unique requirements.In terms of language support, expanding VQA systems to cater to low-resource and underrepresented languages is a pressing need. Leveraging multilingual pretraining, cross-lingual transfer learning, and data augmentation techniques can enable effective question answering in such contexts. For example, the development of AfriBERTa \cite{ogueji-etal-2021-small}, a transformer-based multilingual language model trained on 11 African languages, demonstrates the potential for improving performance in low-resource settings. Addressing code-mixed and dialectal language challenges is equally critical, particularly in linguistically diverse regions like India. The study by \cite{10.1145/3695766} offers a comprehensive survey on NLP tasks, resources, and techniques for Telugu-English code-mixed text, shedding light on strategies to tackle linguistic diversity. Furthermore, recent advancements in multilingual and code-mixed VQA systems, such as the work by \cite{khan2021developingmultilingualcodemixedvisual}, utilize knowledge distillation to enhance system performance in these challenging settings. These efforts highlight the growing recognition of code-mixed and multilingual scenarios in VQA research. Algorithmically, advancements in Vision-Language Models (VLMs) hold promise for improving VQA systems. Future research can focus on modular and lightweight architectures for scalable and efficient models, especially for deployment on edge devices. Enhancing multimodal alignment, temporal reasoning, and spatial understanding will be key to tackling complex video-based queries. The work by \cite{zou2024videodistilllanguageawarevisiondistillation} introduces VideoDistill, a framework with language-aware behavior in both vision perception and answer generation processes, closely resembling human behavior.Emerging research should also address multi-task learning for simultaneous performance across related tasks, such as video captioning and question answering. Efforts to bridge the gap between symbolic reasoning and neural approaches can lead to more holistic and context-aware VQA systems. Furthermore, ethical considerations, including bias mitigation in datasets and equitable multilingual support, are critical for the responsible development of VQA models. The study by \cite{alam2024maya} introduces Maya, an open-source multilingual multimodal model, aiming to handle low-resource languages and varied cultural contexts. Lastly, explainability in VQA systems remains an important goal, with models needing to provide interpretable rationales for their outputs, especially in high-stakes domains like healthcare and legal analysis. The work by \cite{wang2023vaquitaenhancingalignmentllmassisted} introduces VaQuitA, a framework designed to refine the synergy between video and textual information, enhancing alignment in LLM-assisted video understanding.

These directions collectively offer a roadmap for advancing the VQA field, paving the way for broader applicability and deeper understanding of video-based question answering.
\section{Conclusion}\label{sec: Section_9}
This survey provides a comprehensive entry point for those starting with Video Question Answering (VQA), offering a clear narrative of its evolution. By exploring key advancements in representation learning, attention mechanisms, and multimodal alignment, we outlined the critical developments that have shaped VQA systems. The distinction between extractive and abstractive approaches—focusing on direct answer retrieval versus generating more contextual responses—serves as a framework for understanding the versatility of VQA models across different domains. A detailed summary of key models and techniques is presented in Table  ~\ref{tab: summary}, which highlights the contributions of various works. As VQA continues to evolve, this survey highlights the importance of balancing accuracy, efficiency, and reasoning capabilities. For newcomers to the field, this review offers both foundational insights and a roadmap to navigating the ongoing challenges and future opportunities in VQA, equipping them to contribute to this dynamic area of research.

\bibliography{custom}
\bibliographystyle{acl_natbib}

\onecolumn

\begin{landscape}
\centering
\begin{longtable}{|p{5cm}|p{1cm}|p{2cm}|p{1.5cm}|p{3cm}|p{2cm}|p{8cm}|}
\caption{Summary of Models and Methods} \label{tab: summary} \\
\toprule
Model/Method & Year & Modalities & Paradigm & Core Technique & Datasets Used & Key Contribution \\
\midrule
\endfirsthead
\toprule
Model/Method & Year & Modalities & Paradigm & Core Technique & Datasets Used & Key Contribution \\
\midrule
\endhead
\midrule
\multicolumn{7}{r}{Continued on next page} \\
\midrule
\endfoot
\bottomrule
\endlastfoot
Finding Structure in Time & 1990 & Text & Abstractive & Temporal Structure Analysis & NaN & Explores the representation of time-based structures, influencing models in sequential learning tasks. \\
ImageNet & 2009 & Image & Extractive & CNNs & ImageNet & A large-scale image classification database that has influenced the development of deep learning-based visual models. \\
PASCAL Visual Object Classes (VOC) Challenge & 2010 & Image & Extractive & Object Detection & VOC & A benchmark dataset for object detection and classification, widely used in visual recognition research. \\
SUN Database: Large-Scale Scene Recognition from Abbey to Zoo & 2010 & Image & Extractive & Scene Recognition & SUN Dataset & Introduces the SUN dataset, a large-scale scene recognition dataset, for training and evaluating scene classification models. \\
ImageNet Classification with Deep CNNs & 2012 & Image & Extractive & Deep CNNs & ImageNet & Introduces deep convolutional neural networks (CNNs) for image classification, revolutionizing computer vision. \\
Efficient Estimation of Word Representations & 2013 & Text & Extractive & Word2Vec & NaN & Introduces Word2Vec for efficient word representation learning, foundational for many NLP tasks. \\
Very Deep Convolutional Networks for Large-Scale Image Recognition & 2014 & Image & Extractive & Convolutional Networks & ImageNet & Introduces VGG, a deep CNN model, for large-scale image recognition, significantly improving performance in image classification. \\
Rich Feature Hierarchies & 2014 & Image & Extractive & Deep CNNs & ImageNet & Proposes a rich feature hierarchy model for object detection and segmentation using CNNs. \\
Multimodal Learning with Deep CNNs & 2014 & Image+Text & Abstractive & Deep CNNs & NaN & Combines deep CNNs with multimodal data to enhance learning for visual question answering. \\
Microsoft COCO: Common Objects in Context & 2014 & Image+Text & Extractive & Object Detection, Captioning & COCO & A comprehensive dataset for object detection and image captioning, establishing a standard for vision-language tasks. \\
Show and Tell: A Neural Image Caption Generator & 2015 & Image+Text & Abstractive & Image Captioning & MSCOCO & Proposes Show and Tell for automatic image captioning using a deep neural network. \\
Show, Attend and Tell: Neural Image Caption Generation with Visual Attention & 2015 & Image+Text & Abstractive & Visual Attention & MSCOCO & Introduces the attention mechanism for generating image captions, improving caption quality by focusing on relevant parts of the image. \\
Deep Visual-Semantic Alignments & 2015 & Image+Text & Abstractive & Visual-Semantic Alignment & MSCOCO & Introduces a deep model for generating image descriptions based on visual-semantic alignment. \\
VQA: Visual Question Answering & 2015 & Image+Text & Extractive & Convolutional Networks & MSCOCO, Visual7W & One of the first datasets and benchmarks for visual question answering using CNNs. \\
Multilingual Image Question Answering & 2015 & Image+Text & Extractive & Multilingual Models & NaN & Proposes a multilingual framework for answering image-based questions across languages. \\
Stacked Attention Networks for Image Question Answering & 2016 & Image+Text & Abstractive & Attention Networks & VQA & Proposes Stacked Attention Networks for visual question answering to improve reasoning over image content. \\
Multimodal Compact Bilinear Pooling & 2016 & Image+Text & Abstractive & Compact Bilinear Pooling & VQA, MSCOCO & Introduces compact bilinear pooling for efficient multimodal fusion in VQA and visual grounding. \\
Neural Module Networks (NMNs) & 2016 & Image+Text & Abstractive & Modular Neural Networks & VQA & Introduces a modular approach to VQA, where each module handles specific aspects of the question. \\
Hierarchical Memory Networks & 2016 & Text+Image & Abstractive & Memory Networks & VQA & A hierarchical memory network model that addresses complex reasoning in visual question answering. \\
MovieQA: Understanding Stories in Movies through Question-Answering & 2016 & Video+Text & Abstractive & Movie QA & MovieQA Dataset & Proposes MovieQA, a multimodal QA task for understanding stories in movies using video and text data. \\
Bilinear Attention Networks & 2017 & Image+Text & Abstractive & Bilinear Attention & NaN & Introduces bilinear attention for effective multimodal learning in vision-language tasks. \\
CLEVR: Compositional Language and Elementary Visual Reasoning & 2017 & Image+Text & Abstractive & Compositional Reasoning & CLEVR & Provides a diagnostic dataset for compositional language and elementary visual reasoning tasks. \\
Visual Genome & 2017 & Image+Text & Abstractive & Crowdsourced Dense Annotations & Visual Genome & Connects language and vision using dense image annotations, creating a comprehensive dataset for multimodal learning. \\
Graph Neural Networks for Visual Question Answering & 2017 & Image+Text & Abstractive & Graph Neural Networks & VQA & Introduces the use of graph neural networks for VQA tasks to model visual relationships. \\
Inferring Scene Structure & 2017 & Image+Text & Abstractive & Scene Understanding & NaN & Focuses on inferring scene structure and generating narratives from images. \\
MUTAN & 2017 & Image+Text & Abstractive & Tucker Fusion & MSCOCO & Introduces multimodal Tucker decomposition for improved feature fusion in VQA tasks. \\
Visual Question Answering in the Wild & 2017 & Image+Text & Extractive & CNN+LSTM & VQA, MSCOCO & Develops a VQA model that can handle questions in real-world, noisy conditions. \\
Bottom-Up and Top-Down Attention & 2018 & Image & Extractive & Attention Mechanisms & MSCOCO & Integrates bottom-up and top-down attention mechanisms to enhance captioning and VQA performance. \\
Compositional Attention Networks & 2018 & Image+Text & Abstractive & Attention Networks & GQA & Introduces compositional attention networks for machine reasoning, improving VQA. \\
Learning to Answer Visual Questions with Attention on Attention & 2018 & Image+Text & Abstractive & Attention on Attention & NaN & A novel method using attention on attention for VQA, improving question answering by enhancing visual focus. \\
Neural-symbolic Visual Reasoning and Generation & 2018 & Image+Text & Abstractive & Neural-Symbolic Networks & VQA & Introduces neural-symbolic models for visual reasoning and generation in VQA tasks. \\
Dynamic Scene Graph for Visual Question Answering & 2018 & Image+Text & Abstractive & Scene Graphs & VQA & Proposes dynamic scene graphs to improve reasoning about object relationships in images for visual question answering. \\
FigureQA & 2018 & Image+Text & Abstractive & Visual Reasoning & FigureQA & A dataset and model for visual reasoning over annotated figures in academic texts. \\
VizWiz Grand Challenge & 2018 & Image+Text & Extractive & Image-Text Matching & VizWiz & Focuses on answering questions from blind people by combining visual and textual modalities. \\
Visual Question Answering in Radiology & 2018 & Image+Text & Extractive & Image-Text Matching & VQA-RAD & Focuses on visual question answering in radiology images, providing a dataset of clinically generated questions and answers. \\
ImageCLEF 2018 Medical Domain VQA & 2018 & Image+Text & Extractive & Medical Image Understanding & ImageCLEF & Focuses on VQA in the medical domain, emphasizing medical imagery and question answering. \\
VisualBERT: A Simple and Performant Visual Language Model & 2019 & Image+Text & Abstractive & BERT & MSCOCO & Introduces VisualBERT, combining vision and language in a single transformer architecture for multimodal reasoning. \\
VisualBERT: A Simple and Performant Visual Language Model & 2019 & Image+Text & Abstractive & BERT & MSCOCO & Introduces VisualBERT, combining vision and language in a transformer architecture, simplifying multimodal reasoning. \\
Deep Modular Co-Attention Networks for Visual Question Answering & 2019 & Image+Text & Abstractive & Co-Attention Networks & VQA & Proposes deep modular co-attention networks to improve visual question answering by jointly attending to image and question features. \\
LXMERT: Learning Cross-Modality Encoder Representations from Transformers & 2019 & Image+Text & Abstractive & Cross-Modality Transformers & NaN & Introduces LXMERT for cross-modal representation learning using transformers for vision and language tasks. \\
A Corpus for Reasoning About Natural Language Grounded in Photographs & 2019 & Image+Text & Abstractive & Natural Language Reasoning & Photography Dataset & Introduces a corpus for reasoning over natural language grounded in images, focused on real-world scenarios. \\
Social-IQ: A Question Answering Benchmark for Artificial Social Intelligence & 2019 & Image+Text & Abstractive & Social Intelligence & SocialIQ Dataset & Introduces a VQA benchmark focusing on social intelligence to understand human interactions in images. \\
Towards VQA Models That Can Read & 2019 & Image+Text & Abstractive & Vision-Language Pretraining & Visual Question Answering Datasets & Proposes a model that incorporates reading comprehension for better visual question answering. \\
GQA: Real-World Visual Reasoning & 2019 & Image+Text & Abstractive & Visual Reasoning & GQA & A dataset and model designed for real-world visual reasoning and compositional question answering. \\
VideoBERT: A Joint Model for Video and Language Representation Learning & 2019 & Video+Text & Abstractive & Video-Language Pretraining & NaN & Proposes VideoBERT for learning joint representations of video and text for various downstream tasks. \\
MMFT-BERT & 2020 & Image+Text & Abstractive & Multimodal Fusion Transformer & VQA, MSCOCO & A multimodal fusion transformer with BERT encodings for improved VQA performance. \\
OSCAR: Object-Semantics Aware Pre-training & 2020 & Image+Text & Abstractive & Object-Semantic Pre-training & COCO & Introduces OSCAR, which integrates object semantics in pre-training for better vision-language understanding. \\
PlotQA: Reasoning over Scientific Plots & 2020 & Image+Text & Abstractive & Plot Reasoning & Scientific Plot Dataset & Introduces PlotQA, focusing on reasoning over scientific plots in a multimodal context. \\
UNITER & 2020 & Image+Text & Extractive & Representation Learning & MSCOCO, Visual Genome & A unified representation learning model for images and texts, enhancing multimodal understanding. \\
MMBERT: Multimodal BERT Pretraining & 2021 & Image+Text & Abstractive & BERT Pretraining & Medical VQA Datasets & Proposes MMBERT for improving medical VQA by fine-tuning on multimodal data, enhancing domain-specific understanding. \\
Multilingual and Code-Mixed VQA System & 2021 & Image+Text & Abstractive & Knowledge Distillation & NaN & Develops a multilingual, code-mixed VQA system by leveraging knowledge distillation for performance enhancement. \\
Learning Transferable Visual Models from Natural Language Supervision & 2021 & Image+Text & Abstractive & Transferable Visual Models & NaN & Proposes CLIP for learning visual models using natural language supervision, achieving state-of-the-art performance in vision tasks. \\
ViLT: Vision-and-Language Transformer & 2021 & Image+Text & Abstractive & Transformer & NaN & Proposes ViLT, a vision-and-language transformer without convolution or region supervision, for efficient multimodal learning. \\
Small Data? No Problem! Exploring Pretrained Multilingual Models & 2021 & Text & Extractive & Multilingual Pretraining & Multilingual Datasets & Explores the effectiveness of pretrained multilingual language models for low-resource languages. \\
FashionVQA: A Domain-Specific Visual Question Answering System & 2022 & Image+Text & Abstractive & Domain-Specific VQA & Fashion VQA Dataset & Introduces FashionVQA, a domain-specific VQA model tailored for fashion-related images and questions. \\
Flamingo & 2022 & Image+Text & Abstractive & Few-shot Learning Visual Language Model & NaN & Focuses on visual language tasks, leveraging few-shot learning for visual and textual understanding. \\
Knowledge-Augmented Neural Networks for Visual Question Answering & 2022 & Image+Text & Abstractive & Knowledge-Augmented NN & NaN & Proposes knowledge-augmented neural networks for VQA, enhancing reasoning with external knowledge. \\
Multi-Modal Understanding and Generation for Medical Images and Text & 2022 & Image+Text & Abstractive & Vision-Language Pre-training & Medical Image Datasets & Proposes a vision-language pre-training method for medical images and texts to enhance medical understanding. \\
(2.5+1)D Spatio-Temporal Scene Graphs & 2022 & Video & Extractive & Scene Graphs & NaN & A model for video question answering that uses spatio-temporal scene graphs to model interactions over time. \\
Frozen in Time & 2022 & Video+Image & Extractive & Joint Encoder & Kinetics-700, MSCOCO & Proposes a unified model for video and image retrieval, focusing on efficient video-image representation learning. \\
VideoQA: A Comprehensive Survey of Datasets and Methods & 2022 & Video+Text & Extractive & Video Question Answering & VideoQA Datasets & Provides a comprehensive survey of video question answering datasets and methods. \\
BLIP-2: Bootstrapping Language-Image Pre-training & 2023 & Image+Text & Abstractive & Language-Image Pre-training & NaN & Proposes BLIP-2, leveraging frozen image encoders and large language models for vision-language pre-training. \\
Vision–Language Model for VQA in Medical Imagery & 2023 & Image+Text & Extractive & Multimodal Fusion & Medical Image Datasets & A model designed for medical image question answering by using vision-language fusion to handle medical images and related queries. \\
VaQuitA: Enhancing Alignment in LLM-Assisted Video Understanding & 2023 & Video+Text & Abstractive & Video-Language Alignment & NaN & Proposes VaQuitA, which focuses on improving the alignment between large language models and video understanding. \\
GPT-4V’s Multimodal Capability for Medical Image Analysis & 2024 & Image+Text & Abstractive & Multimodal Large Language Models & Medical Image Datasets & Evaluates GPT-4V’s performance on medical image analysis, providing insights on its multimodal capabilities. \\
Maya & 2024 & Text+Image+Video & Abstractive & Instruction Fine-tuned Multilingual Multimodal Model & NaN & A multilingual model fine-tuned with instructions to handle multimodal inputs, improving cross-modal understanding and few-shot performance. \\
Sports-QA & 2024 & Video & Extractive & Video Question Answering & Sports-QA & A large-scale video question answering benchmark for professional sports, enabling complex reasoning in sports videos. \\
VideoDistill: Language-aware Vision Distillation for Video Question Answering & 2024 & Video+Text & Abstractive & Vision Distillation & NaN & Proposes VideoDistill for distilling vision and language models for better video question answering. 
\end{longtable}
\end{landscape}
\end{document}